\documentclass[conference]{IEEEtran}
\IEEEoverridecommandlockouts

\usepackage{cite}
\usepackage{amsmath,amssymb,amsfonts}
\usepackage{graphicx}
\usepackage{textcomp}
\usepackage{xcolor}
\usepackage{multirow}
\usepackage{booktabs}%
\usepackage{amsthm}%
\usepackage{mathrsfs}%
\usepackage{manyfoot}%
\usepackage{algorithm}%
\usepackage{algorithmicx}%
\usepackage{algpseudocode}%
\usepackage{listings}%
\usepackage{subcaption}%

\def\BibTeX{{\rm B\kern-.05em{\sc i\kern-.025em b}\kern-.08em
    T\kern-.1667em\lower.7ex\hbox{E}\kern-.125emX}}
\begin{document}

\title{Risk Stratification for ICU Delirium\\using Pervasive Ambient Sensing Information
\thanks{A.B. and P.R. were supported by NIH/NINDS R01 NS120924, NIH/NIBIB R01 EB029699. P.R. was also supported by NSF CAREER 1750192.}
}

\DeclareRobustCommand*{\IEEEauthorrefmark}[1]{%
  \raisebox{0pt}[0pt][0pt]{\textsuperscript{\footnotesize #1}}%
}

\author{
\IEEEauthorblockN{
\begin{tabular}{c}
Jiaqing Zhang\IEEEauthorrefmark{1}$^{*}$,
Sabyasachi Bandyopadhyay\IEEEauthorrefmark{2}$^{*}$,
Miguel Contreras\IEEEauthorrefmark{3},
Jessica Sena\IEEEauthorrefmark{3},\\
Yuanfang Ren\IEEEauthorrefmark{4},
Andrea Davidson\IEEEauthorrefmark{4},
Ziyuan Guan\IEEEauthorrefmark{4},
Tezcan Ozrazgat-Baslanti\IEEEauthorrefmark{4},\\
Subhash Nerella\IEEEauthorrefmark{3},
Azra Bihorac\IEEEauthorrefmark{4},
Parisa Rashidi\IEEEauthorrefmark{3}$^{\dagger}$
\end{tabular}
}

\IEEEauthorblockA{\footnotesize
\begin{minipage}{0.95\textwidth}
\centering
\IEEEauthorrefmark{{1}}Department of Electrical and Computer Engineering, University of Florida, Gainesville, United States\\
\IEEEauthorrefmark{{2}}Department of Medicine, Stanford University, Stanford, United States\\
\IEEEauthorrefmark{{3}}Department of Biomedical Engineering, University of Florida, Gainesville, United States\\
\IEEEauthorrefmark{{4}}Department of Medicine, University of Florida, Gainesville, United States
\end{minipage}
}

\IEEEauthorblockA{\footnotesize
}

\thanks{$^{*}$ Jiaqing Zhang and Sabyasachi Bandyopadhyay contributed equally to this work.}
\thanks{$^{\dagger}$ Parisa Rashidi is the corresponding author.}
}
\maketitle

\begin{abstract}
Delirium is a common and serious complication in the Intensive Care Unit (ICU), associated with increased morbidity, prolonged hospital stays, and higher healthcare costs. Despite its prevalence, early prediction and prevention remain challenging. Environmental factors such as ambient sound and light may influence the onset of delirium, yet they are often overlooked in risk assessments. In this study, we examined whether light intensity and sound pressure levels can independently predict delirium across multiple prediction horizons. We evaluated four efficient sequential neural network models on data collected from 9 ICUs across 309 patients to predict delirium for 10 prediction-window sizes. We reported feature importance and direction of influence using Shapley Additive Explanations analysis. The convolutional model achieved the strongest discrimination, with AUC = 0.80 on sound data and on combined data. Sound features were the dominant predictors overall. Integrating sound with light improved short-term ($<1$ week) prediction, with the combined model assigning the highest risk immediately after the sensing period. These findings suggest that passive ambient sensing, especially sound, can add a clinically meaningful, interpretable signal for delirium risk estimation and offer a practical pathway to enrich multimodal ICU prediction and prevention strategies.
\end{abstract}


\section{Introduction}
Delirium in the Intensive Care Unit (ICU) is a significant cause of morbidity and mortality in critically ill patients \cite{maldonado2017acute}. Delirium is defined as an acute change in awareness and attention that develops over a short period and can be associated with other neurocognitive conditions, such as memory deficits, disorientation, and hallucinations \cite{lv2025artificial}. Studies suggest that delirium happens in 80\% of mechanically ventilated ICU patients \cite{goldberg2020association,salluh2010delirium,zhang2025melon} and accounts for an annual cost of 4-16 billion USD in this patient population \cite{milbrandt2004costs}. Given the significant financial and health burdens of ICU delirium, its diagnosis and treatment remain imperative. 

\begin{figure*}[ht]
\centerline{\includegraphics[width=0.9\textwidth]{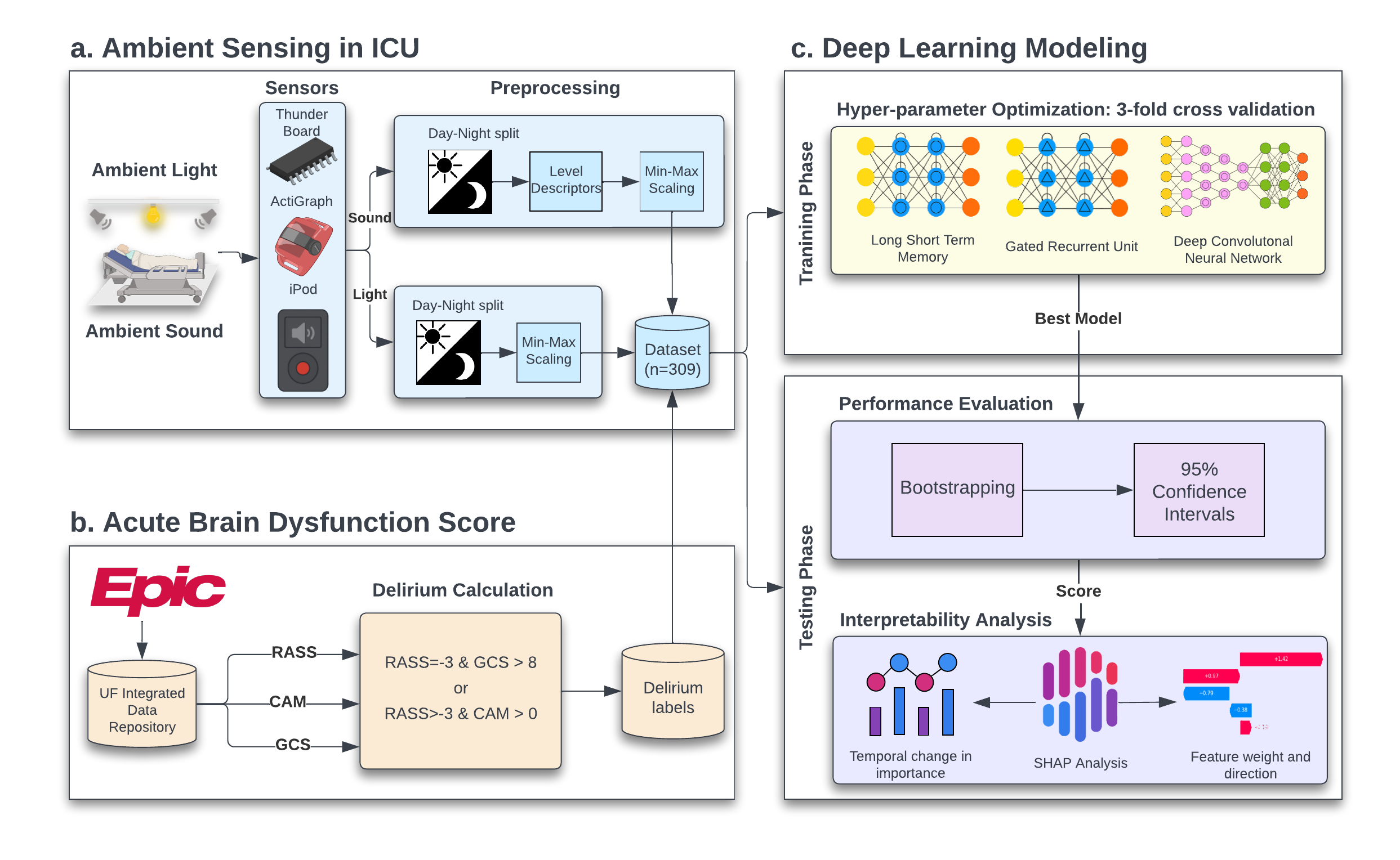}}
\caption{Conceptual Workflow of the methodology. a. Ambient light and sound data from ICU sensors (ThunderBoard, ActiGraph, and iPod) are preprocessed separately for day and night. A dataset (ICU-ENV) of 309 patients is constructed and used to train deep learning models. b. Delirium labels are derived from the electronic health record using RASS, GCS, and CAM scores. c. Three models (LSTM, GRU, and Deep CNN) are evaluated using 3-fold cross-validation. The best-performing model is evaluated on a separate testing set (n = 62) with bootstrapping to estimate 95\% confidence intervals. Interpretability analysis is conducted using SHAP to examine temporal changes in feature importance and direction of contribution.}\label{fig1}
\end{figure*}

Currently, delirium is diagnosed using the Confusion Assessment Method ICU (CAM-ICU) questionnaire \cite{moss2016critical,ely2001evaluation}. The CAM-ICU contains a series of questions attempting to measure the level of attention, organization of thought, consciousness, and change in mental status from baseline \cite{ely2001delirium}. Due to the dependence on a baseline cognitive measurement, CAM-ICU can be inaccurate in patients experiencing post-surgical cognitive complications or baseline neurological disorders \cite{van2011routine}. Furthermore, requiring ICU nurses to administer this test results in sparse measurement, sometimes as low as 38\% of admissions \cite{riekerk2009limitations}. Additionally, CAM-ICU identifies delirium only after it has manifested. Due to the vulnerability of the patient population, identifying the risk of delirium onset can significantly improve the quality of care.

Previous studies have predicted ICU delirium incidence from electronic health records (EHR), vital signs, and EEG signals \cite{ruppert2020icu,gong2023predicting,kim2022machine,sun2019automated}. With the rapid development of large language models (LLMs) used in the clinical domain \cite{zhang2026auditing}, recent work has further explored their ability to detect delirium from EHR \cite{contreras2025dellirium}. Light intensity and sound pressure levels are also found to be a cause of circadian disruption and delirium in the ICU \cite{lee2021association,simons2018noise,ren2025quantifying}, as ICU patients are frequently exposed to prolonged abnormal light and soundscapes \cite{konkani2012noise,lusczek2021light}. Moreover, prior work shows that ICU sound pressure levels are far above the recommended WHO guidelines for hospitals \cite{darbyshire2013investigation}. Although some studies have shown an association between ambient light intensity and sound pressure levels in ICU rooms with delirium development \cite{rodriguez2025impact,sangari2021delirium}, no study has yet developed a delirium risk assessment model using the environmental factors, i.e., sound pressure and light intensity level. 

In this study, we hypothesized that a neural network trained on ambient light intensity and sound pressure levels could prospectively predict which ICU patients are at risk of developing delirium during their ICU stay, as shown in Fig. \ref{fig1}. We used Shapley Additive Explanations (SHAP) analysis to infer the relative importance of the respective features \cite{NIPS2017_7062}.

\section{Materials and Methods}\label{sec2}

\subsection{Participants}

\begin{figure}[h]
\centerline{\includegraphics[width=0.45\textwidth]{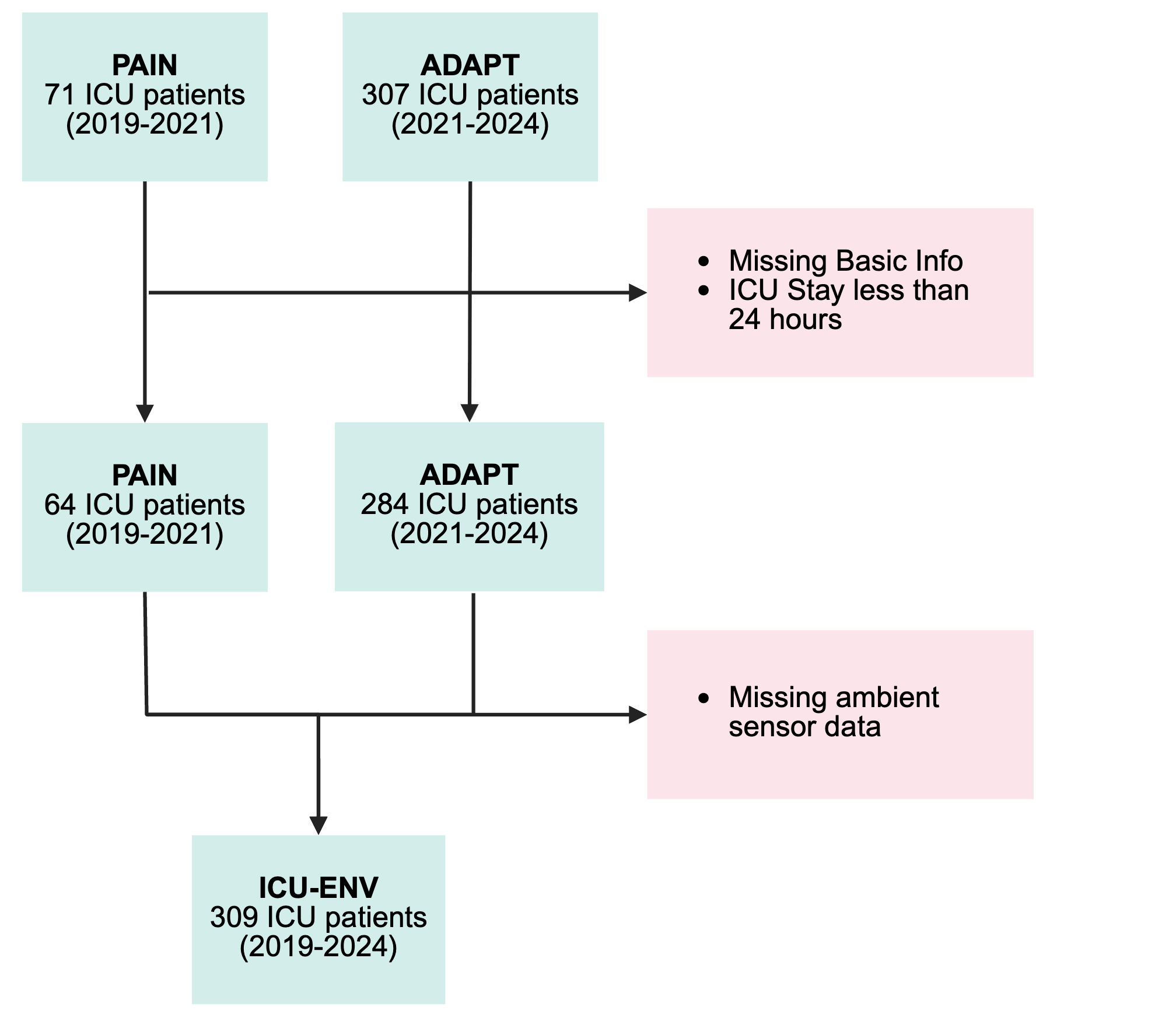}}
\caption{Cohort flow diagram. The ICU-ENV dataset includes 309 ICU patients enrolled between 2019 and 2024, derived from the PAIN and ADAPT studies. Patients were excluded if they had missing basic demographic or clinical information, an ICU stay of less than 24 hours, or incomplete ambient sensor data.}\label{fig2}
\end{figure}

Participants were recruited through two federally funded studies (PAIN and ADAPT) at the University of Florida (UF) Shands Hospital. Informed consent was obtained from either participants or their legally authorized representatives (LARs), who assented on their behalf. Patients were considered eligible if they were older than 18 years old, admitted to a UF ICU, and expected to stay in the ICU for at least 24 hours. Exclusion criteria included discharge, transfer, death within 24 hours of ICU admission, and isolation or contact precaution requirements, as shown in Fig. \ref{fig2}. 

\subsection{Data Collection}

We constructed our ICU-ENV dataset from data collected in two studies between May 2019 and September 2024. Ambient data was collected for seven days or till discharge from the ICU, whichever occurred first. In the first study (PAIN), Actigraph GTX3+ devices (ActiGraph, Pensacola, FL, USA) collected light intensity level data, and iPod Touch 7th generation (Apple Inc., Cupertino, CA, USA) with AudioTools pro web app version 11.2 (Studio Six Digital, Ventura, CA, USA) collected sound pressure level data. In the second study (ADAPT), Thunderboard Sense 2 (Silicon Labs, Austin, TX, USA) collected data on light intensity and sound pressure level. Delirium and coma states were calculated daily using the CAM-ICU, Richmond Agitation Sedation Scale (RASS), and Glasgow Coma Scale (GCS) values according to the algorithm developed by Ren et al. in 2023 \cite{ren2023computable}. According to this algorithm, patients were considered a) comatose if they had at least one RASS score $\leq -3$ and one temporally adjacent GCS score $\leq 8$; b) delirious if they had at least one RASS score equal to $-3$ and one temporally adjacent GCS score greater than 8, or at least one RASS score greater than $-3$ and one temporally adjacent CAM value with “positive” assessment; c) normal if they had at least one RASS score greater than $-3$ and one temporally adjacent CAM value with “negative” assessment. 

\subsection{Preprocessing}

The ICU-ENV dataset comprises two main modalities. The light intensity data (Light dataset) were averaged daily into “daytime” (07:00-18:59) and “nighttime” (19:00-06:59) (according to care providers' shifts), yielding two data points per day (Light-day and Light-night). The sound pressure data (Sound dataset) collected by the Audiotools app contained seven features: maximum sound pressure level (Lmax), minimum sound pressure level (Lmin), a level greater than 99th percentile (L99), 90th percentile (L90), 50th percentile (L50), 10th percentile (L10), and 1st percentile (L01). Accordingly, sound pressure information collected via Thunderboard was converted into the same statistical-level descriptors (Fig. \ref{fig1}) to create a unified dataset. Data from the two studies were scaled between 0 and 1 separately and then combined. The combined dataset was then randomly split into development and test datasets by patient. The same preprocessing pipeline was followed for the Light dataset (except for computing the statistical levels). Finally, the development and test datasets for the Sound and Light dataset were combined to create the Sound + Light dataset. Patients missing either the sound or the light information were removed from the Sound + Light dataset. The features from the three datasets were formatted in sequence by date. Every sequence was zero-padded to 7 days to match the maximum length of sensor data collection, which was defined as the observation window in the study. We defined 10 time points (1-7 days and 14, 21, 28 days) after the observation window as the prediction windows. A patient was considered to be delirious if they experienced at least one delirium episode within the prediction window. 

\begin{figure}[!t]
\centering
\includegraphics[width=0.45\textwidth]{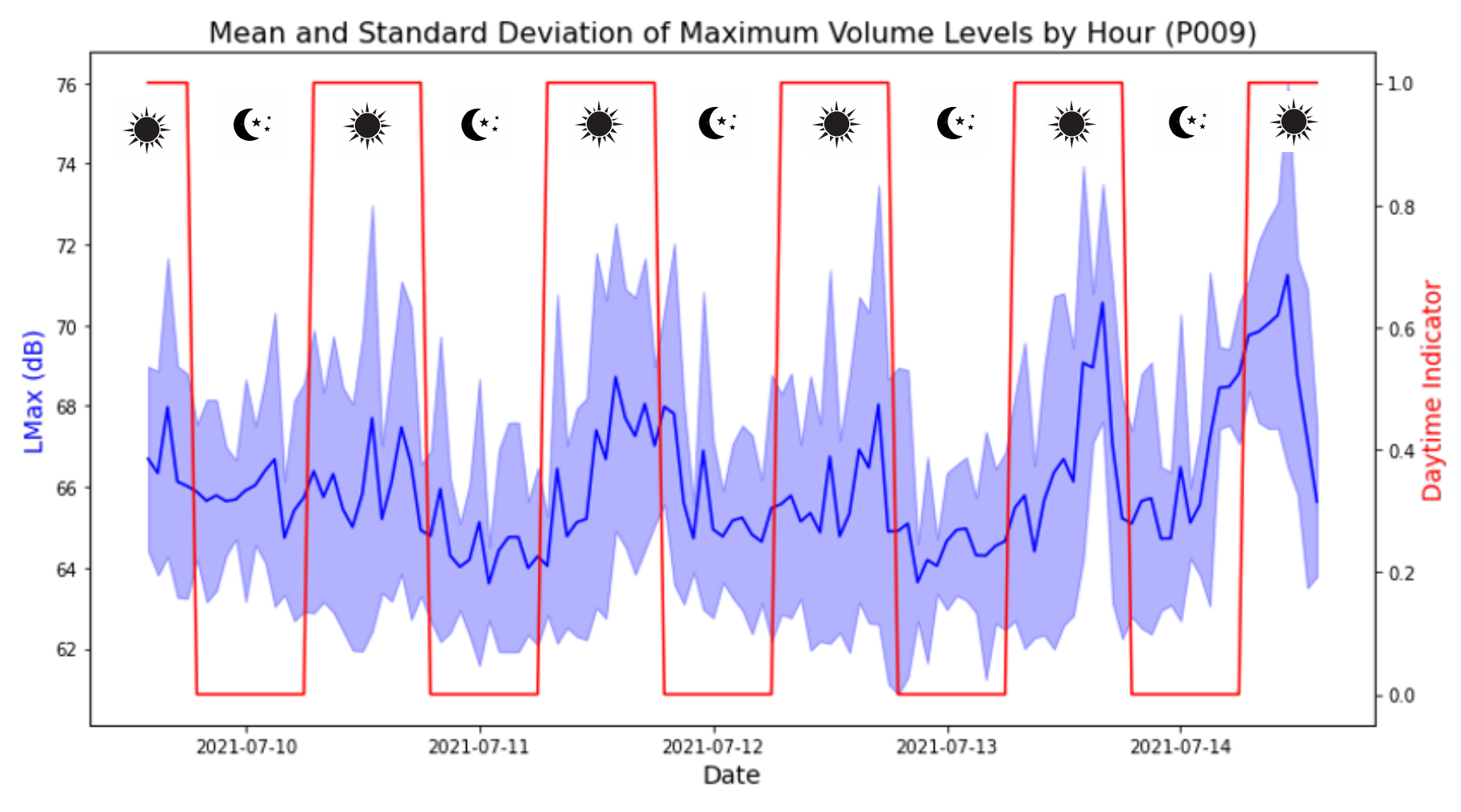}
\caption{Mean and standard deviation of maximum noise in the ICU of a randomly selected patient}
\label{fig3}
\end{figure}

Sound pressure levels in the ICU during daytime and nighttime were significantly different. The maximum daytime versus maximum nighttime sound fluctuations of a randomly selected patient are depicted in Fig. \ref{fig3}.

\subsection{Deep Learning Specifications}

We tested four neural network models: Long Short-Term Memory (LSTM) \cite{hochreiter1997long}, Gated Recurrent Units (GRU) \cite{chung2014empirical}, Convolutional Neural Network (CNN), and Transformer models. Relatively simple sequential models were deemed sufficient as the maximum length of the observation window was seven. All layers except the last fully connected classification layer had the Rectified Linear Unit (ReLU) activation. The classification head had two nodes (delirium vs no-delirium) and a sigmoid activation. The training was conducted with a batch size of 8, a learning rate of 0.001, binary cross-entropy loss, and the Adam optimizer. The best models were selected using the validation sets. The area under the receiver operating curve (AUC), accuracy, F1-score, precision, sensitivity, specificity, and negative predictive value were reported, with 95\% confidence intervals computed from 100 bootstrap samples.

\subsection{Shap Analysis}

 We used SHAP analysis on the best model to identify the most predictive features. For the Sound + Light dataset, SHAP values for sound pressure and light intensity were aggregated separately to investigate the relative weight of each data modality in model performance.

\section{Result}\label{sec3}

\subsection{Participants}

\begin{table}[!t]
\caption{Patient characteristics for the development and test cohorts}\label{tab1}%
\begin{tabular}{@{}llll@{}}
\toprule
Characteristic & Development (N=247) & Test (N=62) & p-value \\
\midrule
Age, mean (SD) & 59.8 (15.6) & 59.2 (17.0) & 0.791 \\
Female, N (\%) & 93 (37.7\%) & 22 (35.5\%) & 0.866 \\
\midrule
\textbf{Race, N (\%)} & & & \\
White & 204 (82.6\%) & 47 (75.8\%) & 0.298 \\
African American & 26 (10.5\%) & 10 (16.1\%) & 0.313 \\
Other & 17 (5.7\%) & 5 (8.1\%) & 0.962 \\
\midrule
\textbf{Comorbidities, N (\%)} & & & \\
Cancer & 18 (7.3\%) & 4 (6.5\%) & 1.000 \\
Cerebrovascular disease & 31 (12.6\%) & 10 (16.1\%) & 0.594 \\
Dementia & 6 (2.4\%) & 3 (4.8\%) & 0.558 \\
Paraplegia/hemiplegia & 20 (8.1\%) & 4 (6.5\%) & 0.867 \\
Peptic ulcer disease & 8 (3.2\%) & 1 (1.6\%) & 0.796 \\
Renal disease & 65 (26.3\%) & 17 (27.4\%) & 0.988 \\
Rheumatologic disease & 11 (4.4\%) & 1 (1.6\%) & 0.505 \\
\toprule
\end{tabular}
\end{table}

\begin{figure}[!t]
\centering
\includegraphics[width=0.45\textwidth]{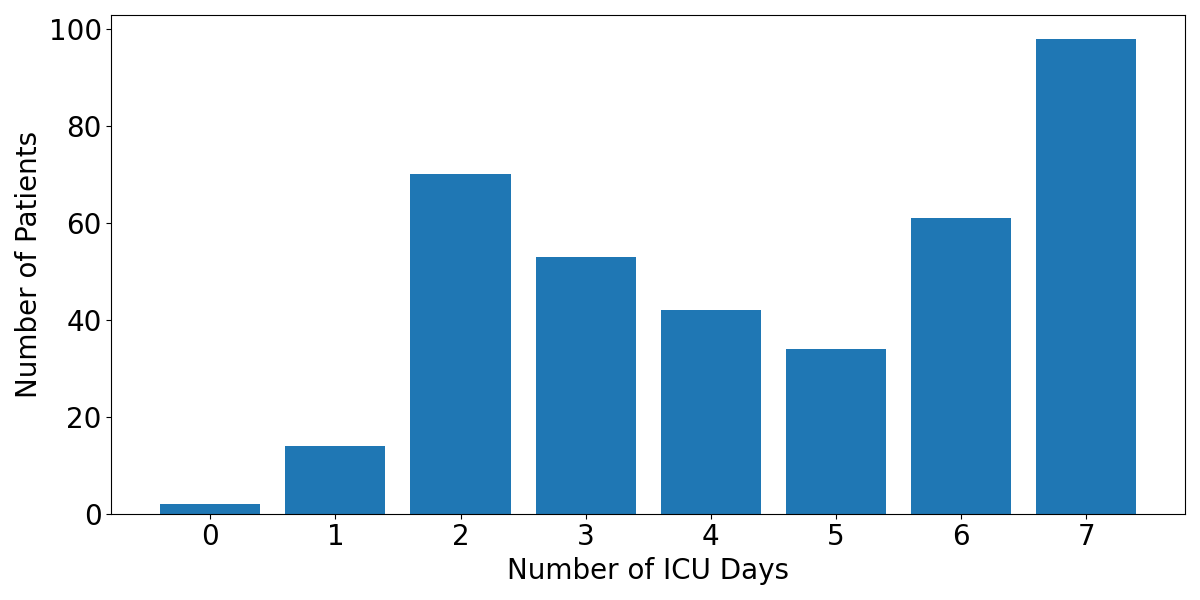}
\caption{Distribution of the number of days of data collection. The distribution is bimodal with peaks at 2 days and 7 days.}\label{fig4}
\end{figure}

The participants in our study were primarily white (11.7\% black, 7.1\% others) and male (37.2\% female), with an average age of 59.7 years (S.D=16.8 years) (Table 1). The average length of data collection was 4.9 days (S.D. = 2.1 days) as shown in Fig. \ref{fig4}. The distribution of the number of days spent in the ICU resembled a bimodal distribution with two peaks at two days and seven days, respectively.

\subsection{Delirium Risk Prediction Using Deep Learning}

\begin{figure}[!t]
\centering
\includegraphics[width=0.5\textwidth]{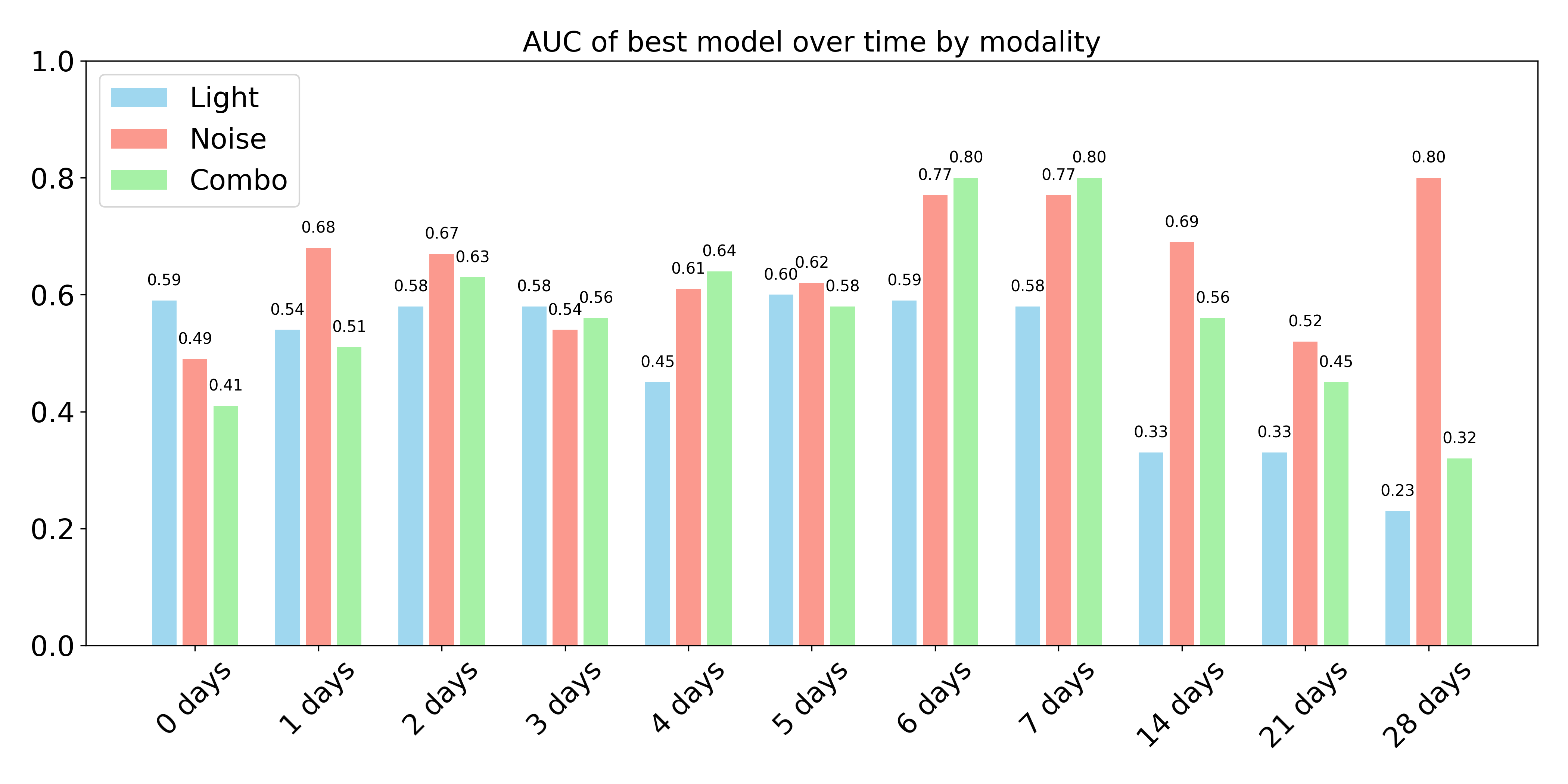}
\caption{Best performances over a 1-month ICU stay for all patients.}\label{fig5}
\end{figure}

The CNN model outperformed the others in all the setups. Across three datasets, the neural network model trained on the Sound dataset achieved better classification performance than one trained on light features. The best classification performance with noise features was in the 28-day prediction window. The model trained on the Light dataset performed best when the prediction window size was 5 days. The model trained on Sound + Light data achieved the highest score when the prediction window size was 7 days. We reported the model performance (CNN only) under different datasets and different prediction window sizes in Fig. \ref{fig5}.

\subsection{Assessment of Delirium Risk Prediction}

\begin{figure}[htbp]
\centering

\begin{subfigure}{0.5\textwidth}
    \centering
    \includegraphics[width=\textwidth]{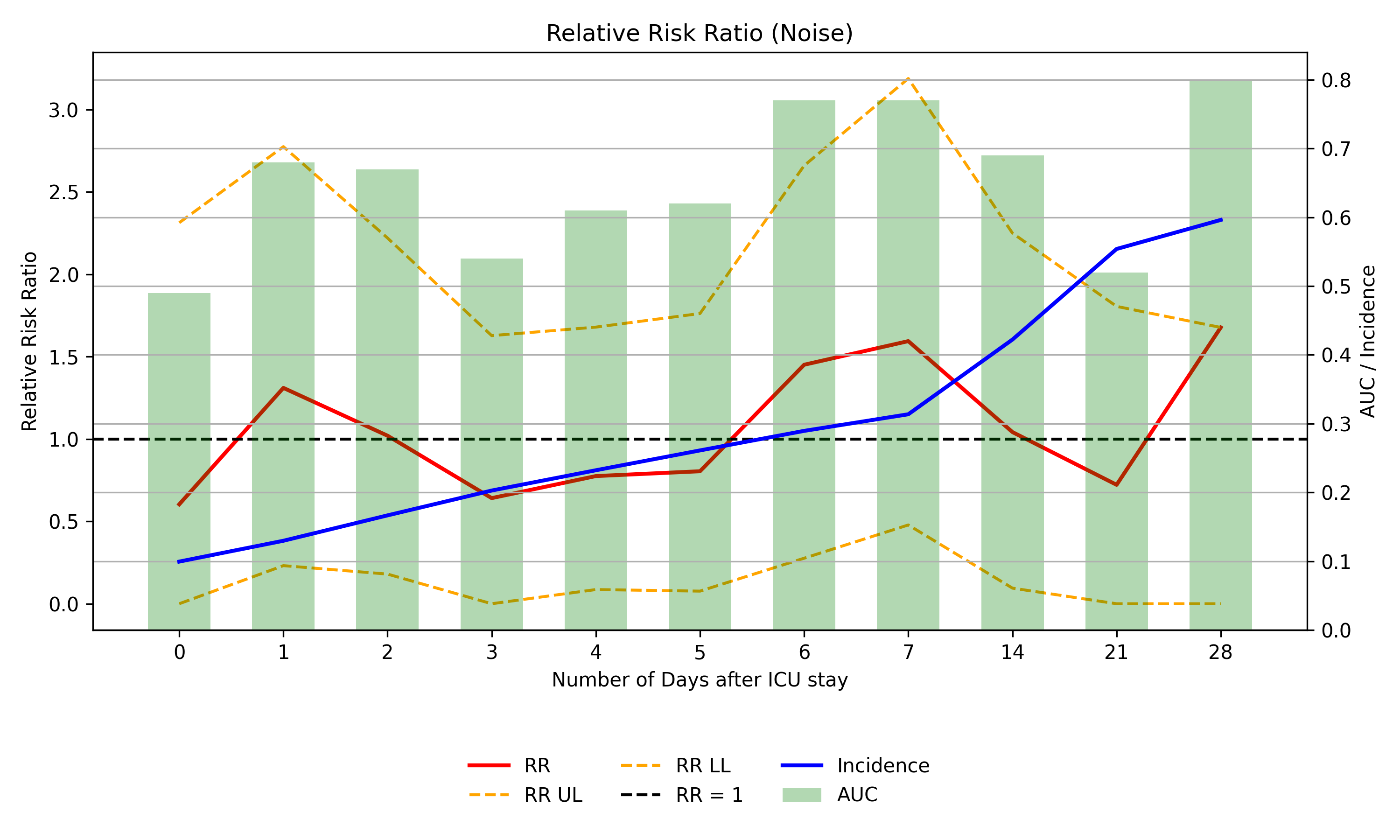}
    \caption{ }
    \label{fig6a}
\end{subfigure}

\begin{subfigure}{0.5\textwidth}
    \centering
    \includegraphics[width=\textwidth]{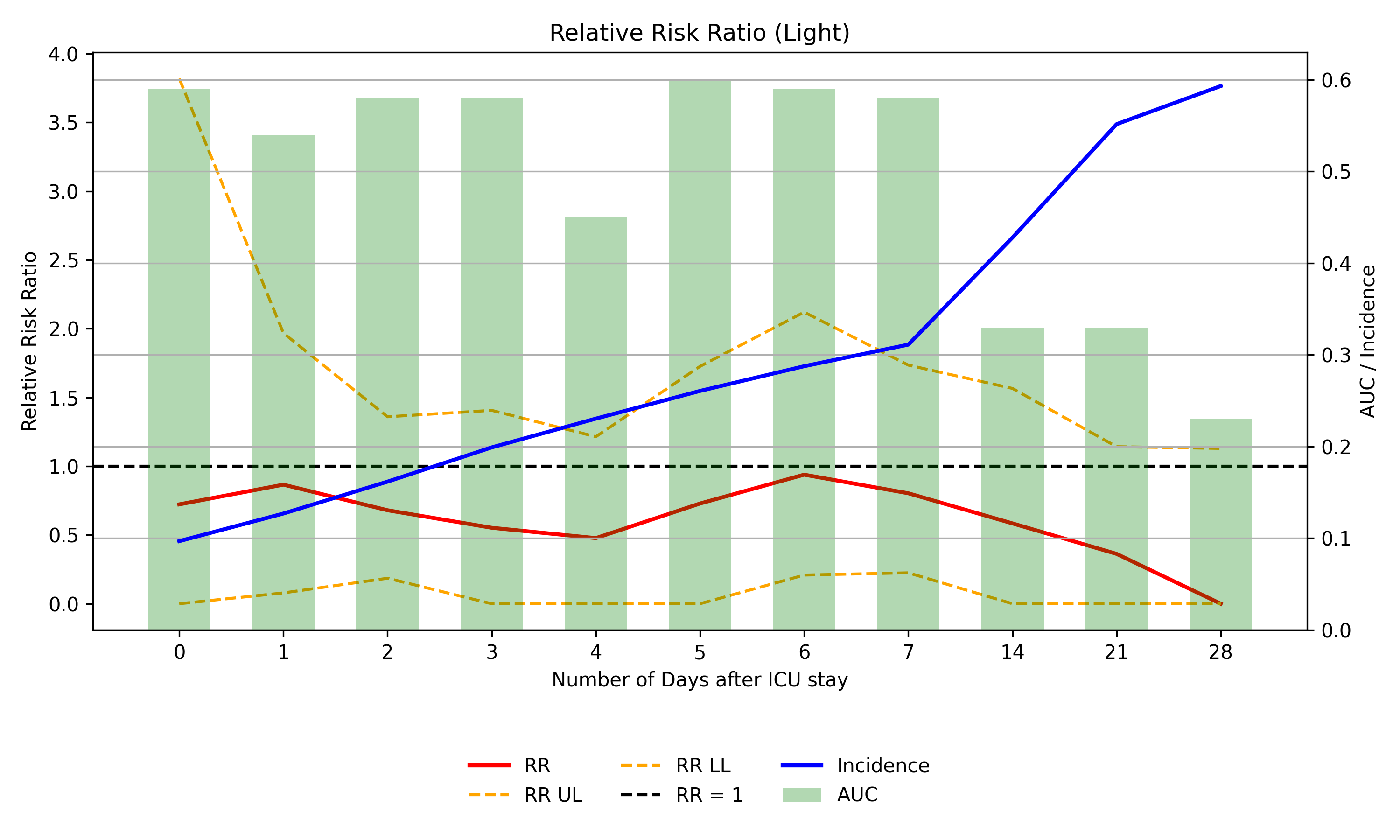}
    \caption{ }
    \label{fig6b}
\end{subfigure}

\begin{subfigure}{0.5\textwidth}
    \centering
    \includegraphics[width=\textwidth]{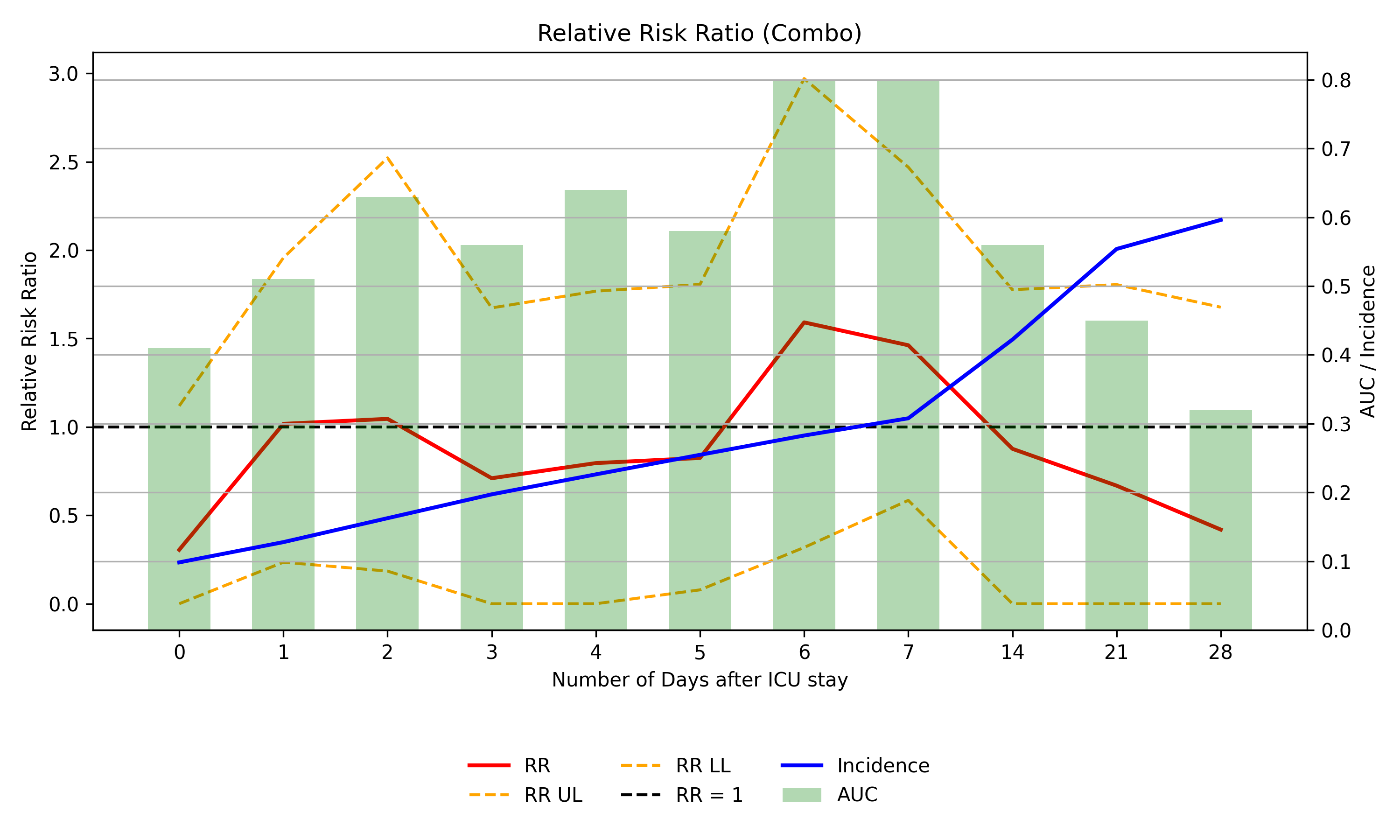}
    \caption{ }
    \label{fig6c}
\end{subfigure}

\caption{(a). Relative risk ratios were calculated for models using only sound pressure data (Sound dataset). (b). Relative risk ratios were calculated for models using only light data (Light dataset). (c). Relative risk ratios were calculated for models using combined data on light and sound pressure.}\label{fig6}
\end{figure}

The ratio between precision and incidence (relative Risk Ratio; RR) was calculated at different time points. Fig. \ref{fig6} shows that the relative RR predicted by sound pressure data briefly increases on the third day after data collection (best model performance coincides with the highest risk) and decreases before rising at 28 days. In contrast, the model trained on light intensity data exhibits an oscillating RR value with peaks at 4 (best performance and highest risk), 7, and 28 days. In comparison, the model trained on combined modalities predicts the highest RR value on the first day after data collection, higher values between three and four days, and lower values otherwise. The maximum RR value using Sound + Light data was 1.8 times that from sound data and 2.6 times that from light data. Thus, although the performance dropped after combining light and sound information, patients were assigned a higher risk of developing delirium at an earlier time point after their sensory data were collected.

\subsection{SHAP Interpretation of Deep Learning Models}

\begin{figure}[htbp]
\begin{subfigure}{0.45\textwidth}
    \centering
    \includegraphics[width=\textwidth]{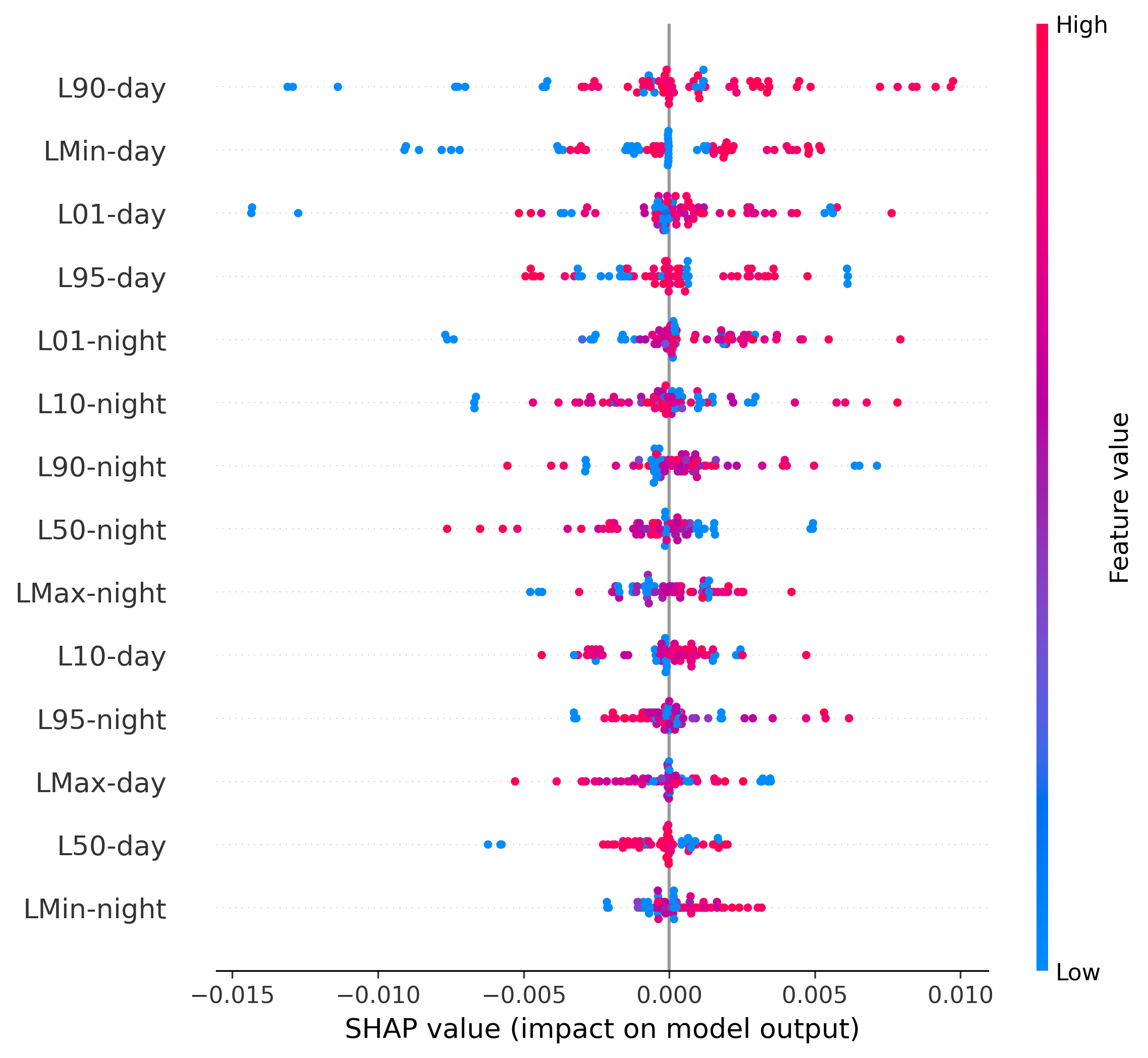}
    \caption{ }
    \label{fig7a}
\end{subfigure}

\begin{subfigure}{0.45\textwidth}
    \centering
    \includegraphics[width=\textwidth]{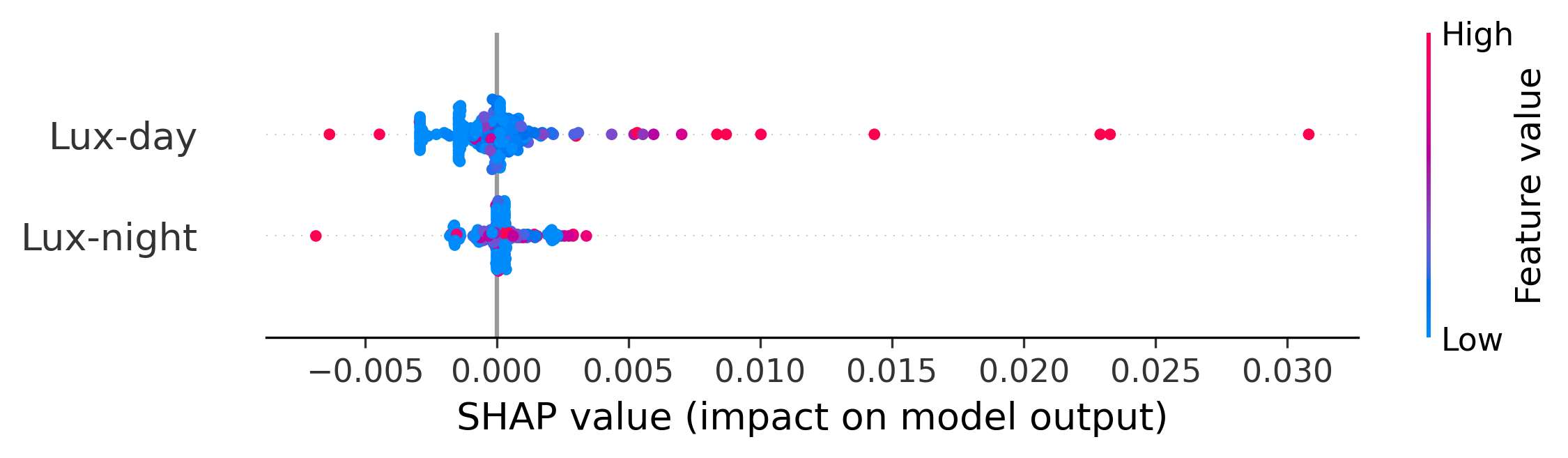}
    \caption{ }
    \label{fig7b}
\end{subfigure}

\begin{subfigure}{0.45\textwidth}
    \centering
    \includegraphics[width=\textwidth]{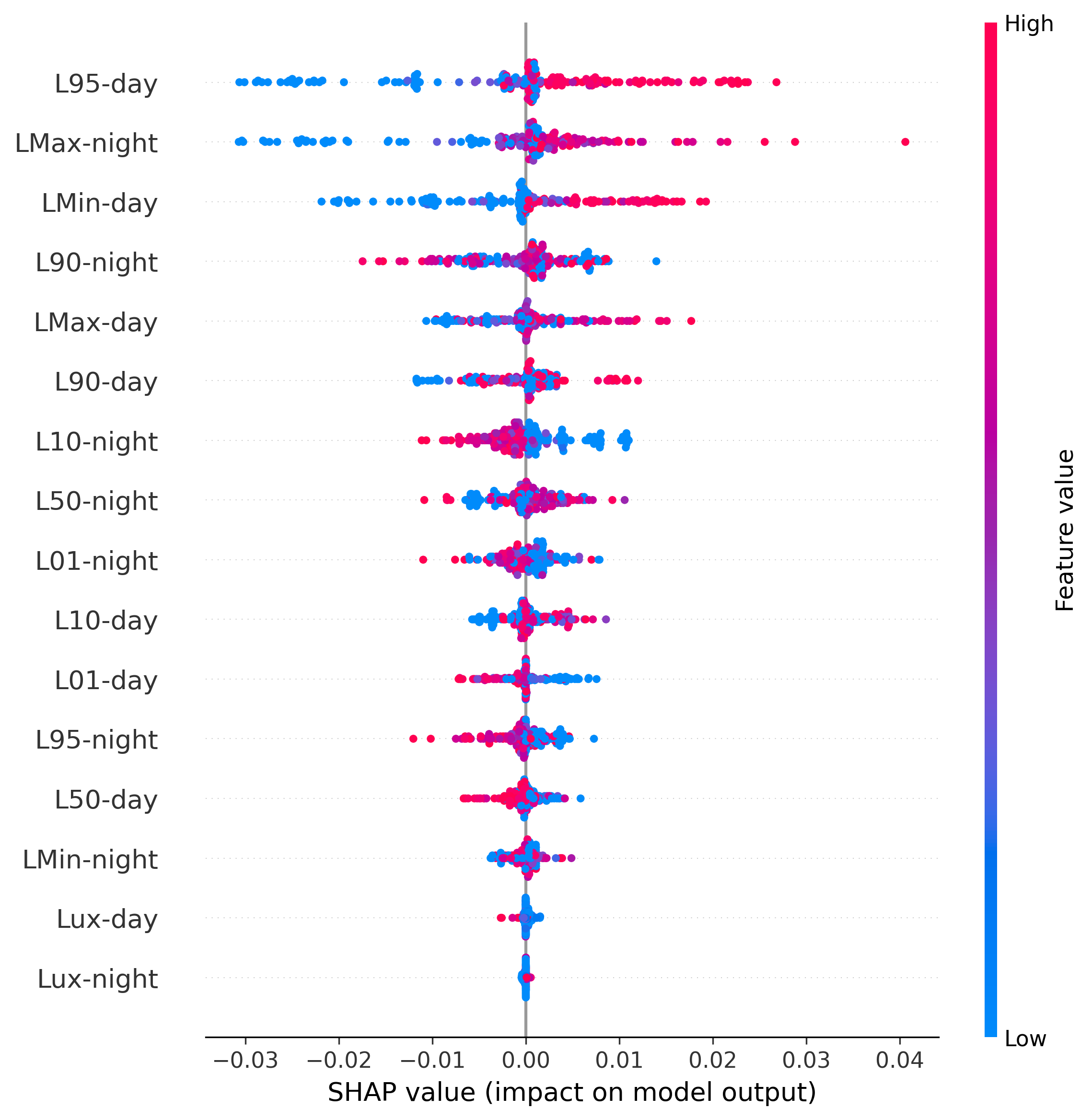}
    \caption{ }
    \label{fig7c}
\end{subfigure}
\caption{(a). Best model with sound (28-day prediction window), (b). best model with light (5-day prediction window), (c). A model that predicted the highest risk with Sound + Light data (7-day prediction). Red dots indicate samples with high values of that particular feature, and blue dots indicate samples with low values of that particular feature. Positive X-axis values indicate that the corresponding feature value pushed the model towards predicting 1; and vice versa.}
\label{fig7}
\end{figure}

The feature importance analyzed by SHAP for the best-performing models is shown in Fig. \ref{fig7}. The L90-day and L01-day sound were the most significant positive and negative predictors of delirium, respectively (Fig. \ref{fig7a}). Correspondingly, their difference is a positive predictor of delirium. This feature's importance indicates that during the daytime, lower background sound (Lmin-day) combined with sustained high foreground sound (L90-day) positively predicted delirium. These were followed by L01-night, which positively predicted the development of delirium. L95-day and L01-night were the most important day and night sound pressure levels, respectively, underscoring the importance of sustained foreground sound during the daytime and intermittent background sound at nighttime. The SHAP analysis revealed that daytime light intensity was more important than nighttime light intensity (Fig. \ref{fig7b}). The sound features were more significant than the light features when training the model on the Sound + Light dataset (Fig. \ref{fig7c}). L95-day and LMax-night were the top 2 predictors. 

\begin{figure}[!t]
\begin{subfigure}{0.45\textwidth}
    \centering
    \includegraphics[width=\textwidth]{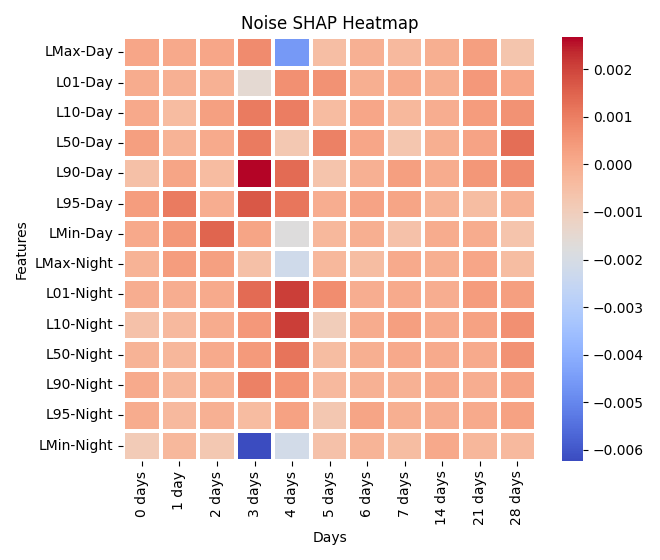}
    \caption{ }
    \label{fig8a}
\end{subfigure}

\begin{subfigure}{0.45\textwidth}
    \centering
    \includegraphics[width=\textwidth]{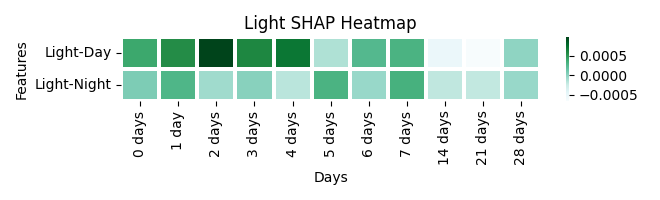}
    \caption{ }
    \label{fig8b}
\end{subfigure}

\caption{(a). shows the SHAP coefficients for the noise cohort and (b). shows that for the light cohort}
\label{fig8}
\end{figure}

We demonstrated the variegated impact of sound and light intensity features across different days of data collection for the best model trained on both modalities (for an observation window of 7 days). The SHAP analysis highlighted the coefficients for the best models from sound and light datasets for ten different time points over one month (Fig. \ref{fig8}). These findings revealed that while all daytime sound pressure levels, except Lmin-day, positively predicted delirium, only Lmax-night was a consistent positive predictor.

\section{Discussion \& Conclusion}

In this study, we used data from two prospective, single-center ICU studies to train three temporal neural network models to predict the risk of delirium over 28 days from the day of data collection, using ambient ICU light and sound data. The models developed using ICU Sound and Sound + Light information showed strong performance in classifying patients with delirium. Models developed using only the Light information showed moderate performance. SHAP analysis revealed that all daytime sound pressure level features except Lmin-day positively predicted delirium. In contrast, among the nighttime sound pressure levels, only the Lmax-night consistently predicted delirium positively. The Sound + Light dataset improved risk prediction while slightly compromising classification performance. SHAP analysis of the best model on the Sound + Light dataset indicated that sound predictors are more informative than light predictors for predicting delirium. 

To our knowledge, this study is the first to investigate the feasibility of predicting ICU delirium incidence and stratifying ICU delirium risk using only ambient factors. Therefore, this study contributes to a fledgling body of research utilizing pervasive sensing in the ICU. Neural network-enabled non-linear modeling of multimodal data. For example, in the best model on the Sound + Light dataset, LMAX-night and L95-day significantly predicted delirium, with no linear relationship to the class label, as shown by the SHAP plots. SHAP analysis revealed the relative contribution of sound and light features across the observation window. 

\begin{figure}[!t]
\begin{subfigure}{0.24\textwidth}
    \centering
    \includegraphics[width=\textwidth]{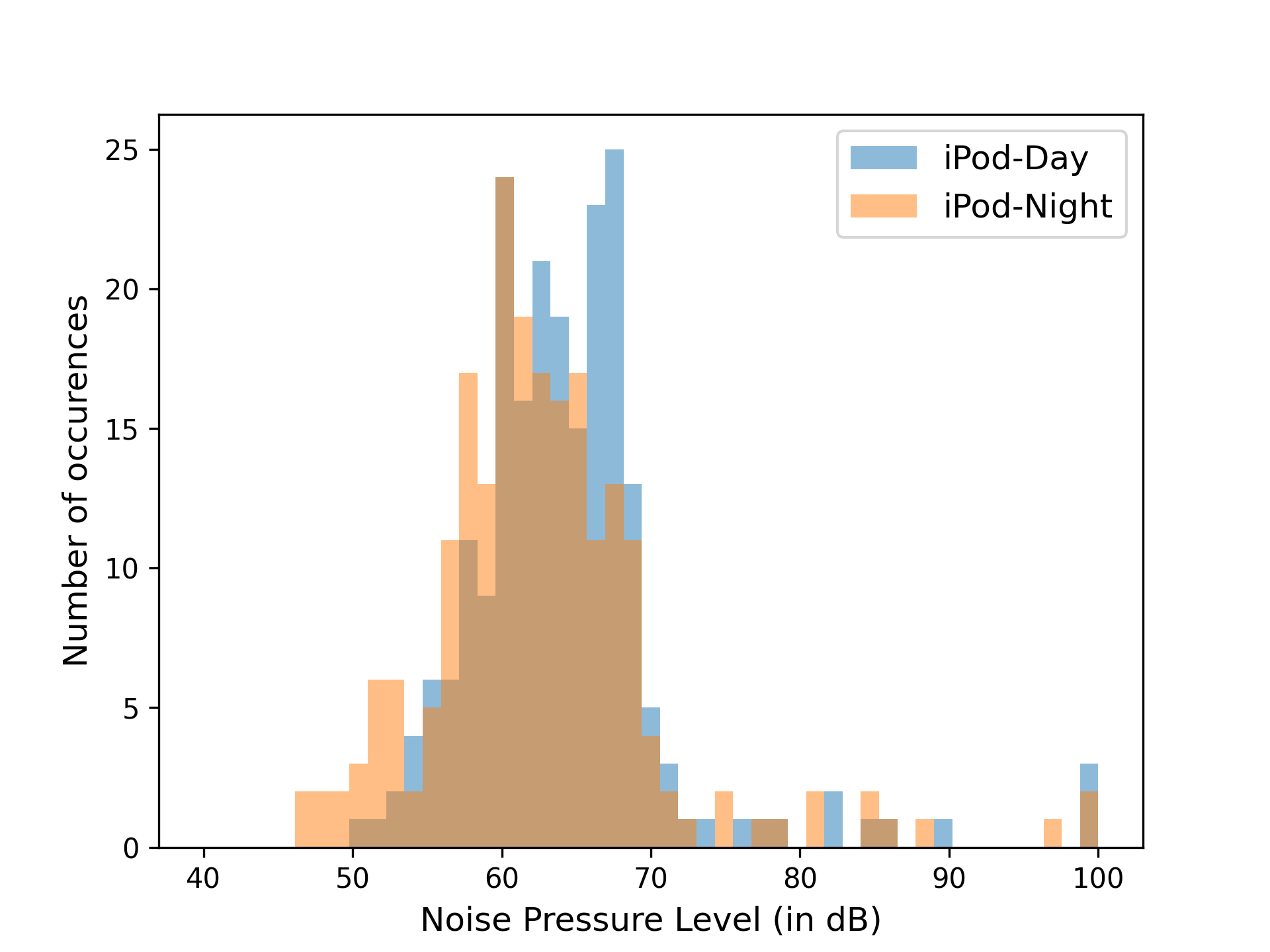}
    \caption{ }
    \label{fig9a}
\end{subfigure}
\hfill
\begin{subfigure}{0.24\textwidth}
    \centering
    \includegraphics[width=\textwidth]{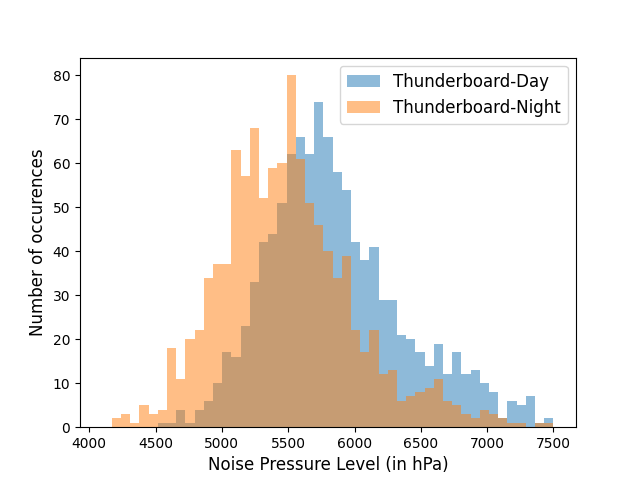}
    \caption{ }
    \label{fig9b}
\end{subfigure}

\begin{subfigure}{0.24\textwidth}
    \centering
    \includegraphics[width=\textwidth]{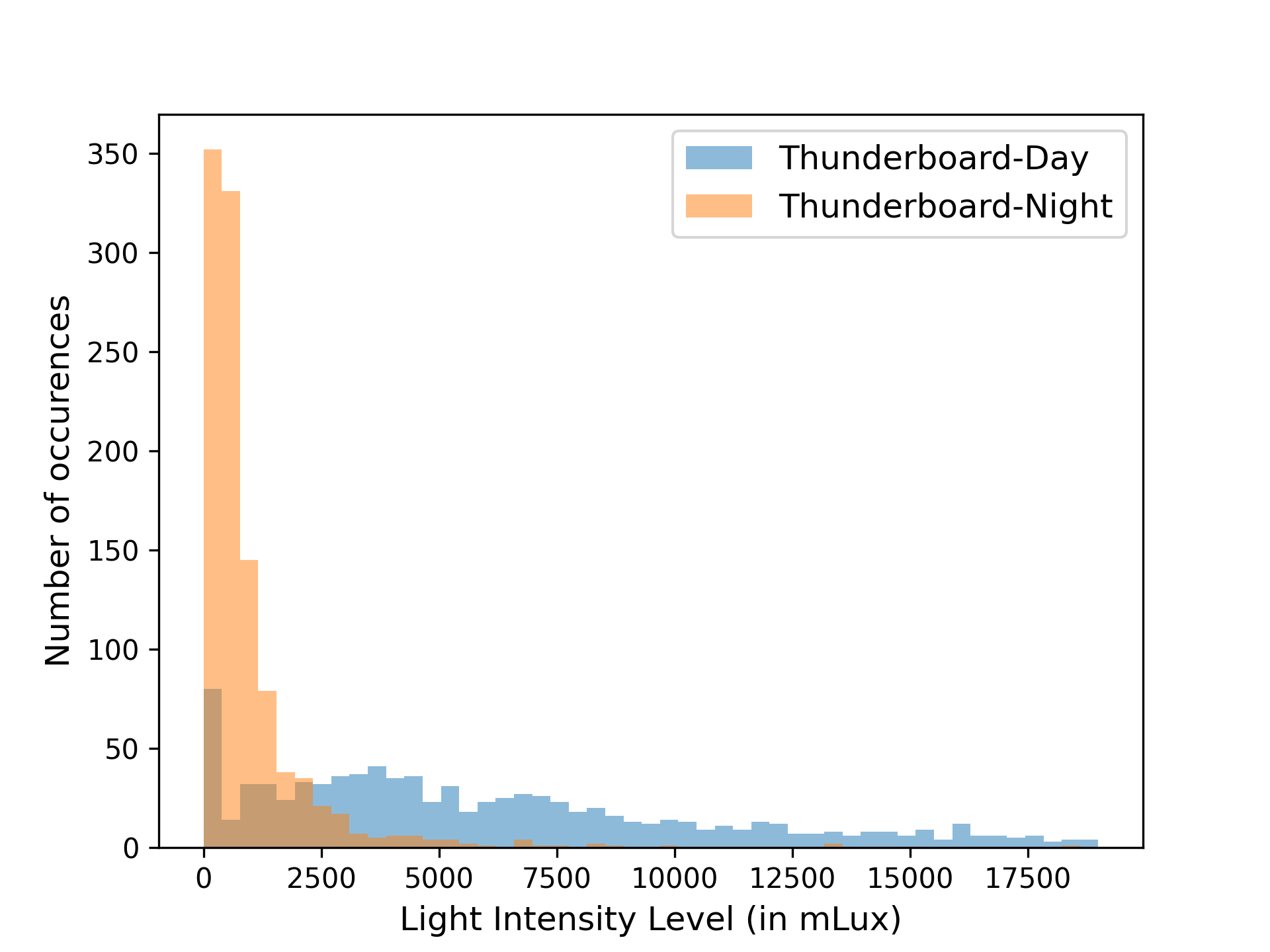}
    \caption{ }
    \label{fig9c}
\end{subfigure}
\hfill
\begin{subfigure}{0.24\textwidth}
    \centering
    \includegraphics[width=\textwidth]{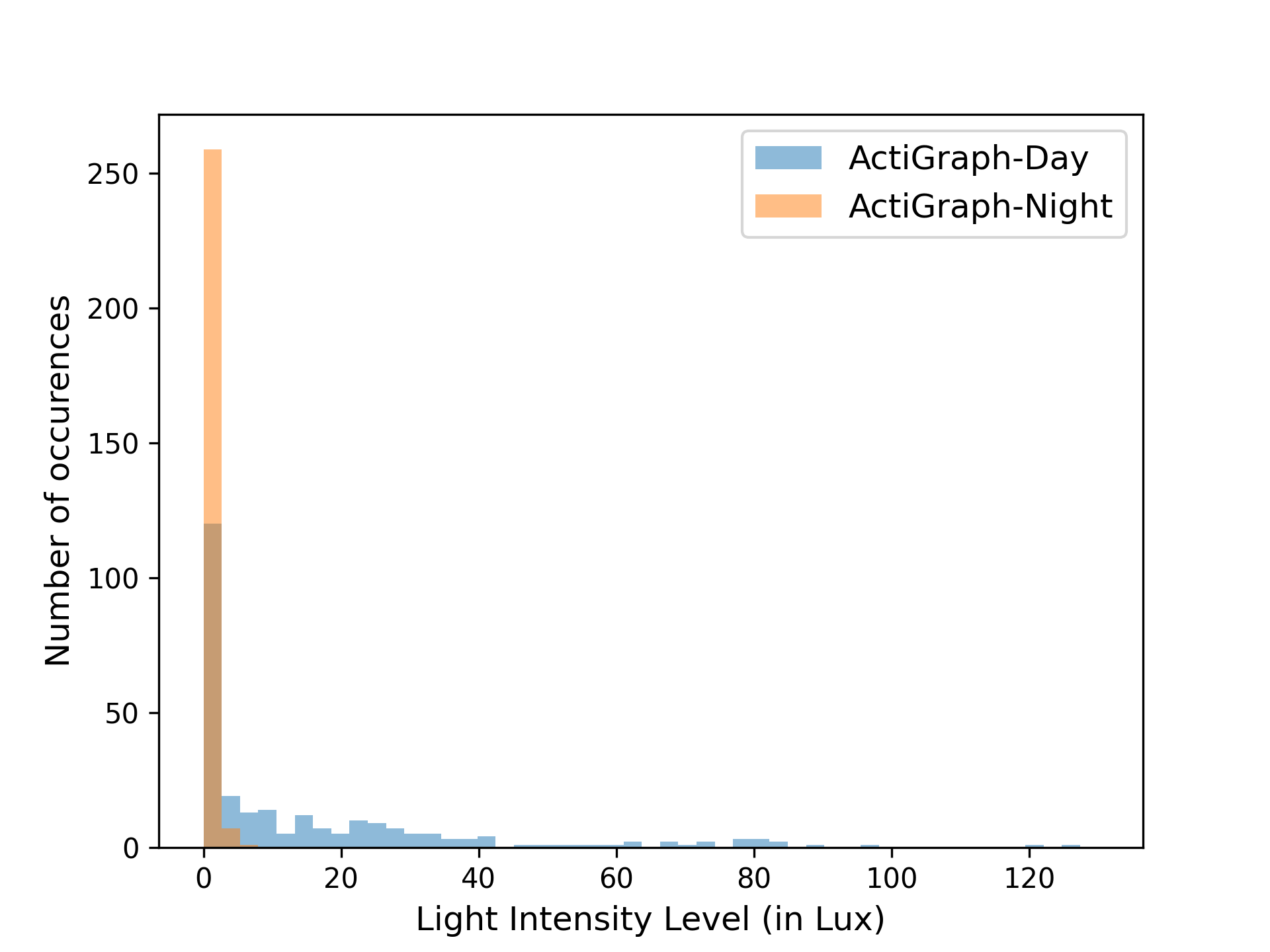}
    \caption{ }
    \label{fig9d}
\end{subfigure}

\caption{(a). Noise levels collected through the iPod. (b). Noise levels collected through Thunderboard. (c).  Light levels collected using ActiGraph. (d). Light levels collected using Thunderboard. Noise intensities are Gaussian across sensors, whereas light intensities are positively skewed and differ between the ActiGraph and Thunderboard sensors.}
\label{fig9}
\end{figure}

The presence of 309 patients in our overall cohort compromised the robustness of the classification models, necessitating further data collection. This smaller sample size prevented us from training models with a large number of parameters, as it would have increased the risk of overfitting. Another disadvantage of using multiple sensors was the possible presence of batch effect in the data. In the sound data, we addressed this by computing the statistical-level descriptors reported by the AudioTools app using the Thunderboard data. The sound pressure distributions in both studies were Gaussian (Fig. \ref{fig9a} and \ref{fig9b}). However, the distributions of light intensities collected with the ActiGraph and Thunderboard were different (Fig. \ref{fig9c} and \ref{fig9d}). This could be a factor contributing to the loss of performance when using light intensities alone. 

Future work will expand data collection to include richer environmental signals and other complementary modalities, enabling more robust multimodal training \cite{zhang2024mango}. We will further refine the model and conduct a prospective evaluation and deployment in a real-world ICU setting.

\section*{Glossary of Terms}
CNN: convolutional neural network; AUC: area under the receiver operating curve; CAM-ICU: Confusion Assessment Method ICU; EHR: electronic health records; GCS: Glasgow Coma Scale; ICU: Intensive Care Unit; GRU: Gated Recurrent Units; LAR: legally authorized representatives; Lmax: maximum sound pressure level; Lmin: minimum sound pressure level; L99: a level greater than 99th percentile; L90: a level greater than 90th percentile; L50: a level greater than 50th percentile; L10: a level greater than 10th percentile; L01: a level greater than 1st percentile; LSTM: Long Short Term Memory; RASS: Richmond Agitation Sedation Scale; ReLU: Rectified Linear Unit; RR: relative Risk; S.D: Standard Deviation; UF: University of Florida.

\bibliographystyle{IEEEtran}
\bibliography{bibliography}

\end{document}